# Clickbait in Hindi News Media : A Preliminary Study

**Vivek Kaushal**
International Institute of
Information Technology, Hyderabad
India
vivek.kaushal@research.iiit.ac.in

**Kavita Vemuri**
International Institute of
Information Technology, Hyderabad
India
kvemuri@iiit.ac.in

## Abstract

A corpus of Hindi news headlines shared on Twitter was created by collecting tweets of 5 mainstream Hindi news sources for a period of 4 months. 7 independent annotators were recruited to mark the 20 most retweeted news posts by each of the 5 news sources on its clickbait nature. The clickbait score hence generated was assessed for its correlation with interactions on the platform (retweets, favorites, reader replies), tweet word count, and normalized POS (part-of-speech) tag counts in tweets. A positive correlation was observed between readers' interactions with tweets and tweets' clickbait score. Significant correlations were also observed for POS tag counts and clickbait score. The prevalence of clickbait in mainstream Hindi news media was found to be similar to its prevalence in English news media. We hope that our observations would provide a platform for discussions on clickbait in mainstream Hindi news media.

## 1 Introduction

A news headline provides a brief introduction to the news story and perhaps more importantly, lays emphasis on the focus and scope of the accompanying news article. A common journalistic advice is to present a clear attention-grabbing headline but to not exaggerate and misreport the news story or mislead the reader. With the advent of social media and news aggregators, newsreaders are encountering headlines from a variety of sources – established traditional publishers, up-and-coming online news sources, individual writers, among others – some of which apply questionable tactics to attract the attention of readers. This bombardment of information has led to an overload and hence readers gravitate towards sensationalism as a valid filter (Molek-Kozakowska, 2013).

In the digital media space, catchy headlines that lure readers to click on them and link to accompanying articles are called 'clickbait' (Chakraborty et al., 2016; Chen et al., 2015b). The social media site Facebook has defined clickbait as an implementation of forward referencing – "when a publisher posts a link with a headline that encourages people to click to see more, without telling them much information about what they will see"(El-Arini and Tang, 2014). There is a growing fear that the line between traditional headlines, predatory clickbait and fake news is rapidly blurring (Chen et al., 2015a). While there is agreement that not all catchy headlines are malicious, provided the corresponding article satisfies the information gap in the headline with factual reporting (Rony et al., 2017), the concern raised in journalistic circles is the intent to misuse literary techniques and mislead readers for mere profit. Additionally, one needs to look into whether the traffic generated by clickbait headlines with minimal or untrue information leads to social distrust and affects cognitive processes. That is, are clickbait akin in nature to rumors and half-truths and more importantly, what is the impact of headline-only-readers (as reported by the satirical news site Science Post, which posted a catchy headline with nonsensical article content and found that nearly 46,000 shared the article – without even opening the article linked to the headline)[1] who could propagate misinformation by filling in the knowledge gaps by unverified intuitions and information from uncertified sources.

The existing body of research on clickbait is overwhelmingly focused on the English language, except a few notable studies (Orosa et al., 2017; Gabielkov et al., 2016). This includes research on clickbait's reach, impact and its detection. In this preliminary study, we extend the English language clickbait headlines analysis to Hindi. The interest is to understand whether journalists from non-

---

[1]http://thesciencepost.com/study-70-of-facebook-commenters-only-read-the-headline/

English papers apply clickbait as strategy, as was found in a study of 28 EU news sources (Orosa et al., 2017). By also considering the sharing behavior of headline readers, the aim is to understand relations between the demand-supply of clickbait headlines.

We have reported the findings of a pilot study conducted on news headlines shared by five mainstream Hindi news sources on Twitter. In the country where the study was conducted, approximately 40% of the population's mother tongue is Hindi[2]. All the recruited annotators had Hindi as their native tongue and English as a secondary language. The large and rapidly growing reader-base which is increasingly present on social media, places the relevance of this study.

We have annotated the most retweeted news headlines of the five selected news sources on its clickbait-nature and have done POS tagging of tweets' text. Through a detailed correlation analysis of the 'clickbait-score' generated through manual annotation, with tweets' interaction parameters (likes, retweets, replies), text word count and normalized POS tag counts, we hope to understand the existence and reach of clickbait in Hindi news media. The need to understand reader's reading behavior is important for a deeper discussion on the proliferation of clickbait, as studies have shown the impact of clickbait in causing higher stress levels and lowering productivity (Mark, 2014).

## 2 Methodology

### 2.1 Preparing the Corpus

Hindi news publishers with text as their primary medium were selected based on their popularity on social media. The selected news sources included: BBC Hindi, Dainik Jagran, Dainik Bhaskar, Hindustan and Navbharat Times. The tweets from the respective publishers' Twitter handles were scraped for a period of 4 months (May, 2020 till the end of August, 2020). Besides the tweeted text, information was also collected on each tweet's interactions on the platform, quantified by the number of readers' replies, retweets and favorites. The usage of mentions and hashtags in the tweets were also recorded. The scraped tweets were filtered to remove instances of polls, cartoons, images and videos. Tweets that were replies to other users were also excluded from the corpus. After filtering, the generated corpus had around 52,000 tweets of definite news headlines invariably accompanied by a link to a news article on the publisher's website [3]. The corpus was further processed to remove mentions and hashtags used at the end of tweets to promote them on the platform. The few instances where mentions or hashtags were used in running text of headlines were not removed as doing so would have disrupted the semantic structure of the tweet. This corpus was sorted according to the number of retweets and 20 top tweets were selected for each of the 5 news publishers. No particular filter was applied on the topic of the news headline, in this preliminary study.

| News Source | Most Retweeted News |
|---|---|
| BBC Hindi | यूट्यूब पर 'नापसंद' की जा रही है पीएम मोदी के 'मन की बात' |
| Dainik Jagran | ट्विटर पर प्रधानमंत्री के फॉलोअर्स की संख्या छह करोड़ पहुंचना उनकी बढ़ती लोकप्रियता का प्रमाण है? |
| Hindustan | #TonyKakkar और #ShehnaazGill का नया गाना कुर्ता पजामा रिलीज हो गया है। |
| Navbharat Times | बीजेपी नेता छपवा रहा था NCERT की 'नकली किताबें', 35 करोड़ का माल बरामद |
| Dainik Bhaskar | सोशल मीडिया पर JEE-NEET के खिलाफ कैंपेन: शिक्षामंत्री पर भड़के स्टूडेंट्स बोले- अपनी वेब साइट पर पोल कराकर देख लीजिए, पता चल जाएगा कितने बच्चे परीक्षा देना चाहते हैं |

Table 1: Most retweeted news item for each news source in our dataset.

### 2.2 Clickbait Annotation

7 independent bi-lingual annotators (age: $\mu = 31.6$ years, $\sigma = 13.5$ years) with Hindi as their native tongue and English as a secondary language were provided with standard English definitions and English examples of clickbait and non-clickbait headlines. The annotators were provided with only the Devanagari script text of the 100 most retweeted Hindi news headlines presented in a random order and asked to rate each of them on a 5-point linear scale – [0, 0.25, 0.5, 0.75, 1] where 0 implies not clickbait at all and 1 implies strongly clickbait. The annotation setup was similar to existing research on English news tweets (Potthast et al., 2018), as standard definitions and examples of clickbait were provided to the annotators before they assigned a clickbait-score to presented headlines based on their own perception. The responses were saved and mean of all the 7 annotators' response was taken to generate a 'clickbait score' for each tweet.

---

[2] https://www.thehindu.com/data/what-percentage-of-people-prefer-to-speak-hindi-across-states/article27451589.ece

[3] The dataset of scraped hindi new media tweets is open and available at https://github.com/kaushalvivek/hindi-media-tweets

## 2.3 Data Analysis

Post annotation, correlation of the generated clickbait score with different parameters of tweets, including the number of replies, retweets, favorites and text word count were studied. A POS tagger based on the Hidden Markov Model was used to assign part of speech tags following the Viterbi algorithm (Ekbal et al., 2007). POS tag counts in tweets were normalized for tweets' word count to isolate the relation between POS tags' occurrence and clickbait score. The correlation between normalized POS tag counts and clickbait score was studied and reported.

The D'Agostino-Pearson Test was conducted on all distributions in the dataset to check for normality, a pre-condition for correlation analysis. All the distributions in our assessment were found to be normal, hence Pearson's test was conducted to evaluate correlation between different variables. The number of annotators were low, but the Cronbach's alpha was calculated for the 7 independent annotators to check for internal consistency and the annotations were found to be consistent ($\alpha \geq 0.8$).

## 3 Results

The average clickbait score for all the annotated Hindi news tweets was $\mu = 0.433$, $\sigma = 0.30$. BBC Hindi had the highest average clickbait score ($\mu = 0.59$, $\sigma = 0.30$), while Dainik Bhaskar had the lowest average clickbait score ($\mu = 0.22$, $\sigma = 0.21$). Clickbait scores of all the news sources in the dataset are illustrated (in figure 1). 21 out of the 100 most retweeted news headlines had a clickbait score $\geq 0.75$, whereas 36 out of the 100 most retweeted news headlines had a clickbait score $\leq 0.25$.

A positive correlation was observed between clickbait score of tweets and all their interaction parameters on Twitter - replies (r = 0.25), retweets (r = 0.19) and favorites (r = 0.18). While the correlation was strongly negative (r = -0.39) between the word count of tweets and their clickbait score. As would make intuitive sense, the interaction parameters in themselves were very strongly correlated with each other ($r \geq 0.7$). The findings are illustrated in table 2.

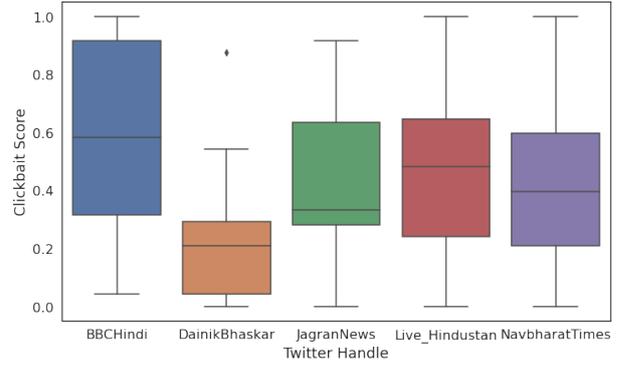

Figure 1: Boxplot of the clickbait score for each news source in our dataset. BBC Hindi ($\mu = 0.59$, $\sigma = 0.30$), Dainik Bhaskar ($\mu = 0.22$, $\sigma = 0.21$), Dainik Jagran ($\mu = 0.44$, $\sigma = 0.27$), Hindustan ($\mu = 0.47$, $\sigma = 0.29$) and Navbharat Times ($\mu = 0.44$, $\sigma = 0.29$).

| Tweets' Parameter | Correlation with Clickbait Score (r) | Significance of Correlation (p) |
|---|---|---|
| Replies | 0.25 | <0.05 |
| Retweets | 0.19 | 0.05 |
| Favorites | 0.18 | 0.07 |
| Word Count | -0.39 | <0.001 |

Table 2: Pearson's r (correlation) between clickbait score and tweets' parameters.

Upon normalizing POS tag counts in each tweet and assessing its correlation with clickbait score, VAUX (r = 0.18), WQ (r = 0.44), NNP (r = 0.17) and QO (r = 0.10) were found to be positively correlated with clickbait score, while INJ (r = -0.18), DEM (r = -0.21) and SYM (r = -0.25) were found to be negatively correlated with clickbait score. No significant correlation was found for other POS tags. The results are illustrated in table 3 in a categorization of POS tags based on correlation with clickbait score.

## 4 Discussion

Against common perception that clickbait is used largely by fringe players who do not post mainstream news material, Rony et al. (2017) had found that 33.54% of social media posts by mainstream English media was clickbait in nature. The ra-tio was even more alarming at 47.56% for mainstream broadcast media. The trend is reflected in our study, where we found that 21% of the most retweeted news headlines by mainstream Hindi print news media had a high clickbait score (clickbait score ≥ 0.75), which is fairly close to the 24.12% clickbait-content ratio for English print news media found by Rony et al. (2017). The similarities between the ratio of clickbait content posted by mainstream English and Hindi news publishers is an indication that proliferation of clickbait is not limited to English. This points to a need to replicate suitable measures for AI-based detection and management of clickbait as are being developed for the English language (Chakraborty et al., 2016; Agrawal, 2016; Biyani et al., 2016; Zhou, 2017; Venneti and Alam, 2018).

| Correlation | POS Tags |
|---|---|
| Positive (>0.1) | VAUX, WQ, NNP, QO |
| Negative (<-0.1) | INJ, DEM, SYM |
| Not Significant | JJ, RP, PRP, INTF, VM, NEG, NN, RB, QF, NST, PSP, CC, XC, QC |

Table 3: Correlation of POS Tags normalized for tweet word-count and clickbait score.

Clickbait score for our Hindi corpus was positively correlated with interaction parameters on Twitter, hence indicating that clickbait content is shared more widely and attracts higher reader attention. This observation is in agreement with existing clickbait research in the English language(Chakraborty et al., 2017) . A strong negative correlation observed between clickbait-score and word-count indicates that clickbait headlines in Hindi are shorter than traditional non-clickbait headline. This is an interesting result that needs to be further examined to isolate effects from language-structure, as similar research studies in the English language have found non-clickbait headlines to be significantly shorter than clickbait headlines (Chakraborty et al., 2016).

The strong positive correlation observed between normalized WQ (question word) counts and clickbait score is expected, as forward-reference – a clickbait technique which involves the omission of a key piece of information in the headline, frequently relies on framing headlines as questions. E.g. "पाकिस्तान में 15 रुपए सस्ता हुआ पेट्रोल, भारत में क्यों नहीं?" – which translates to "Petrol prices reduced by Rs.15 in Pakistan, why not in India?" This headline does not provide any information on why petrol prices are not being reduced in India but generates curiosity by framing the headline as a question, and possibly the emotional/political connotations of the countries being compared. A positive correlation was observed between normalized counts of VAUX (Auxiliary Verb), NNP (Proper Noun), QO (Ordinals) and clickbait score, and a negative correlation was found between INJ (Interjection), DEM (Demonstrative), SYM (Symbol) and clickbait score. These findings require further study and analysis to fully understand the role of each POS tag in evoking curiosity and grabbing attention. Journalistic guidelines suggest that auxiliary verbs are not necessary for perfect passive structures, a proper noun contextualizes and ordinals convey a position or rank which are recommended. Hence, the positive correlation of the two latter POS with clickbait scores is explicable. The negative correlation with interjection is unanticipated as emotional valence attached is considered to be a strong bait for human attention. The similarities between the prevalence of clickbait content in mainstream Hindi news media with recorded observations in English news media is insightful as the research to counter such content in Hindi is fairly primitive. The authors of this paper are working to understand the impact of clickbait on the credibility of news media as perceived by readers (Kaushal and Vemuri, 2020) and plan on scaling their work to news headlines in the Hindi language. We hope that our work would provide a platform for discussion about clickbait in mainstream Hindi news media.

## 5 Limitations

The preliminary study and the inferences drawn is limited by the low number of annotators (7) and the filter applied to study only 100 headlines. Further, cross-reference to article content of a headline was not included in this study, thus limiting the scores from only the headlines. The length of the headlines was not controlled, though not shown to affect the clickbait score, it could weigh the attention score.

## References

A. Agrawal. 2016. Clickbait detection using deep learning. In *2016 2nd International Conference on Next*

*Generation Computing Technologies (NGCT)*, pages 268–272.

Prakhar Biyani, Kostas Tsioutsiouliklis, and John Blackmer. 2016. "8 amazing secrets for getting more clicks": Detecting clickbaits in news streams using article informality. In *Proceedings of the Thirtieth AAAI Conference on Artificial Intelligence*, AAAI'16, page 94–100. AAAI Press.

A. Chakraborty, B. Paranjape, S. Kakarla, and N. Ganguly. 2016. Stop clickbait: Detecting and preventing clickbaits in online news media. In *2016 IEEE/ACM International Conference on Advances in Social Networks Analysis and Mining (ASONAM)*, pages 9–16.

Abhijnan Chakraborty, Rajdeep Sarkar, Ayushi Mrigen, and Niloy Ganguly. 2017. Tabloids in the era of social media? understanding the production and consumption of clickbaits in twitter. *Proc. ACM Hum.-Comput. Interact.*, 1(CSCW).

Yimin Chen, Nadia K. Conroy, and Victoria L. Rubin. 2015a. News in an online world: The need for an "automatic crap detector". *Proceedings of the Association for Information Science and Technology*, 52(1):1–4.

Yimin Chen, Niall J. Conroy, and Victoria L. Rubin. 2015b. Misleading online content: Recognizing clickbait as "false news". In *Proceedings of the 2015 ACM on Workshop on Multimodal Deception Detection*, WMDD '15, page 15–19, New York, NY, USA. Association for Computing Machinery.

Asif Ekbal, Samiran Mandal, and Sivaji Bandyopadhyay. 2007. Pos tagging using hmm and rule-based chunking. *Shallow Parsing for South Asian Languages*, page 25.

Khalid El-Arini and Joyce Tang. 2014. Click-baiting. *About Facebook*.

Maksym Gabielkov, Arthi Ramachandran, Augustin Chaintreau, and Arnaud Legout. 2016. Social clicks: What and who gets read on twitter? In *Proceedings of the 2016 ACM SIGMETRICS International Conference on Measurement and Modeling of Computer Science*, SIGMETRICS '16, page 179–192, New York, NY, USA. Association for Computing Machinery.

Vivek Kaushal and Kavita Vemuri. 2020. Clickbait - trust and credibility of digital news. IEEE International Symposioum on Technology and Society (under review, accepted with changes).

Gloria Mark. 2014. You won't believe what happened next. *The New York Times*.

Katarzyna Molek-Kozakowska. 2013. Towards a pragma-linguistic framework for the study of sensationalism in news headlines. *Discourse & Communication*, 7:173–197.

B Gracia Orosa, S Gallur Santorun, and X Lopez Gracia. 2017. Use of clickbait in the online news media of the 28 eu member countries. *Revista Latina de Comunicación Social*, 72:1.261–1.277.

Martin Potthast, Tim Gollub, Matthias Hagen, and Benno Stein. 2018. The clickbait challenge 2017: Towards a regression model for clickbait strength. *CoRR*, abs/1812.10847.

Md Main Uddin Rony, Naeemul Hassan, and Mohammad Yousuf. 2017. Diving deep into clickbaits: Who use them to what extents in which topics with what effects? In *Proceedings of the 2017 IEEE/ACM International Conference on Advances in Social Networks Analysis and Mining 2017*, ASONAM '17, page 232–239, New York, NY, USA. Association for Computing Machinery.

Lasya Venneti and Aniket Alam. 2018. How curiosity can be modeled for a clickbait detector. *CoRR*, abs/1806.04212.

Yiwei Zhou. 2017. Clickbait detection in tweets using self-attentive network. *CoRR*, abs/1710.05364.